\documentclass[11pt,a4paper]{article}
\usepackage[hyperref]{acl2020}
\usepackage{times}
\usepackage{latexsym}
\usepackage{graphicx} 
\usepackage{natbib}
\usepackage{layout}
\usepackage{tabularx}
\usepackage{amssymb}
\usepackage{amsmath}
\usepackage{amsthm}
\usepackage{booktabs}
\usepackage{algorithm}
\usepackage{algorithmic}
\usepackage{multirow}

\usepackage{microtype}

\aclfinalcopy 

\title{Diformer: Directional Transformer for Neural Machine Translation}

\author{Minghan Wang\textsuperscript{\rm 1}, 
  Jiaxin Guo\textsuperscript{\rm 1}, 
  Yuxia Wang\textsuperscript{\rm 2},
  Daimeng Wei\textsuperscript{\rm 1},
  Hengchao Shang\textsuperscript{\rm 1},
  
  \\
  {\bf Chang Su\textsuperscript{\rm 1},}
  {\bf Yimeng Chen\textsuperscript{\rm 1},}
  {\bf Yinglu Li\textsuperscript{\rm 1},}
  {\bf Min Zhang\textsuperscript{\rm 1},}
  {\bf Shimin Tao\textsuperscript{\rm 1},}
  {\bf Hao Yang\textsuperscript{\rm 1}}
  \\
  \textsuperscript{\rm 1}Huawei Translation Services Center, Beijing, China \\
  \textsuperscript{\rm 2}The University of Melbourne, Melbourne, Australia\\
  \tt \{wangminghan,guojiaxin1,weidaimeng,shanghengchao,\\ 
  \tt suchang8,chenyimeng,liyinglu,zhangmin186,\\
  \tt taoshimin,yanghao30\}@huawei.com \\
  \texttt{yuxiaw@student.unimelb.edu.au}}

\date{}

\begin{document}
\maketitle
\begin{abstract}

Autoregressive (AR) and Non-autoregressive (NAR) models have their own superiority on the performance and latency, combining them into one model may take advantage of both.
Current combination frameworks focus more on the integration of multiple decoding paradigms with a unified generative model, e.g. Masked Language Model.
However, the generalization can be harmful on the performance due to the gap between training objective and inference.
In this paper, we aim to close the gap by preserving the original objective of AR and NAR under a unified framework.
Specifically, we propose the Directional Transformer (Diformer) by jointly modelling AR and NAR into three generation directions (left-to-right, right-to-left and straight) with a newly introduced direction variable, which works by controlling the prediction of each token to have specific dependencies under that direction.
The unification achieved by direction successfully preserves the original dependency assumption used in AR and NAR, retaining both generalization and performance.
Experiments on 4 WMT benchmarks demonstrate that Diformer outperforms current united-modelling works with more than 1.5 BLEU points for both AR and NAR decoding, and is also competitive to the state-of-the-art independent AR and NAR models.
\end{abstract}

\section{Introduction}
\label{sec:intro}
Machine translation can be considered as a conditional generation task, which has been dominated by neural networks, 
especially after Transformer \citep{DBLP:conf/nips/VaswaniSPUJGKP17}.
Conventional autoregressive (AR) NMT models obtain the impressive performance, but it's time-consuming to decode token one by one sequentially \citep{DBLP:conf/nips/SutskeverVL14,DBLP:journals/corr/BahdanauCB14}.
Aiming at fast inference, non-autoregressive (NAR) NMT models enhance the parallelizability by reducing or removing the sequential dependency on the translation preﬁx inside the decoder, but suffering from performance degradation owing to the multi-modality problem, which is still an open-question \citep{DBLP:conf/iclr/Gu0XLS18,DBLP:conf/aaai/ShuLNC20,DBLP:conf/icml/GhazvininejadKZ20,DBLP:conf/emnlp/LeeMC18,DBLP:conf/emnlp/GhazvininejadLL19,DBLP:conf/icml/SternCKU19,DBLP:conf/icml/WelleckBDC19, DBLP:journals/tacl/GuLC19,DBLP:conf/nips/GuWZ19}.

It's always non-trivial to balance high performance and low latency in a single model perfectly.
Therefore, another branch focuses on the \textbf{unified-modeling} of multiple decoding paradigms so that decoding with AR or NAR in different scenarios (AR for quality-first and NAR for speed-first) with one model can be achieved \citep{mansimov2020generalized,DBLP:conf/coling/TianWCLZ20,DBLP:conf/icml/QiG0YCLTLCZ0D21}, making the performance and speed can be pursued more practically.

Whereas, challenges still exist.
For example, a generalized conditional language model is often required to support the generation with customized orders or positions \citep{mansimov2020generalized,DBLP:conf/coling/TianWCLZ20}, which actually prevents the model from being fully trained on specific decoding method, leading to the declines in overall performance.
In addition, in some works, AR and NAR decoding may needs to be trained separately in the stage of pretraining or fine-tuning \citep{DBLP:conf/icml/QiG0YCLTLCZ0D21}, making the training more expensive.

To ameliorate these issues, we propose Directional Transformer (Diformer) which resolve the unification of AR and NAR in a more practical way.
First of all, we abandon the compatible of multiple flexible decoding strategies, but focusing on the modeling of some commonly used strategies that have good performance.
For the AR decoding, it has been proved that monotonic linear generation is still considered as the best strategy \citep{mansimov2020generalized,DBLP:conf/coling/TianWCLZ20}, so we choose to only model the left-to-right (L2R) and right-to-left (R2L) generation.
For the NAR decoding, we choose to follow the stream of masked-language model, like mask-predict in CMLM \citep{DBLP:conf/emnlp/GhazvininejadLL19} or parallel easy-first in Disco \citep{DBLP:conf/icml/KasaiCGG20}, since they are simpler than insertion-based method but still being effective.

To this end, we unify two decoding paradigms into three generation directions --- \textbf{L2R}, \textbf{R2L} and \textbf{straight}, and formulate it through a new objective named as \textbf{Directional Language Model (DLM)}, making the prediction of tokens conditioned on contexts controlled by a newly introduced direction variable.
It ties AR and NAR into a unified generation framework while still preserving the original dependency assumptions of AR and NAR, retaining both generalization and performance.
Meanwhile, all directions can be trained simultaneously with the time spent equally to the training of an independent NAR model, which greatly reduces the training cost compared to two-stages methods.

Experimental results on the WMT14 En$\leftrightarrow$De and WMT16 En$\leftrightarrow$Ro datasets for all three directions indicate that Diformer performs better than previous unification-based works by more than 1.5 BLEU points.
Comparing to other state-of-the-art independent AR and NAR models, Diformer is also competitive when decoding in the same mode.
We summarize contributions of our work as:
\begin{itemize}
    \item We unify the AR and NAR decoding into three generation direction and formulate it with the \textbf{Directional Language Model}.
    \item We propose the Diformer, a Transformer-based model that can be trained with DLM, where all direction can be trained simultaneously.
    \item Experiments on WMT14 En$\leftrightarrow$De and WMT16 En$\leftrightarrow$Ro demonstrate the ability of Diformer with competitive results compared to unified or independent models.
\end{itemize}

\subsection{Related Work}
\cite{mansimov2020generalized} unifies decoding in directed and undirected models by a generalized framework, in which the generating process is factorized as the position selection and the symbol replacement, where the first step is achieved by Gibbs sampling or learned adaptive strategies, the second step can be handled by a masked language model pretrained on monolingual corpora and fine-tuned on the NMT task.
Their model supports at least 5 decoding strategies including hand-crafted and learned, all of them can be used for both linear time decoding (AR) and constant time decoding (NAR).

Similarly, \citet{DBLP:conf/coling/TianWCLZ20} unified AR and NAR by adapting permutation language modeling objective of XLNet to conditional generation, making it possible to generate a sentence in any order.
The model is evaluated to decode in monotonic and non-monotonic AR, semi-AR and NAR with at least 8 position selection strategies including pre-defined and adaptive.

Both of them achieves the compatible to customized decoding through position selection and applying the selected positions/orders on a generalized generative model, which leads to the gap between training and inference.
In contrast to the position selection, we directly model the decoding process with three generation directions in a task-specific manner, thereby without introducing additional complexity to the task and close the gap between training objective and inference strategy.
We consider it is worthwhile to obtain performance improvements by abandon some flexibility.
\section{Method}

\subsection{Background}
Before the description of Diformer, the conventional AR model and the iterative mask prediction based NAR model that applied in Diformer will be introduced first.

The likelihood of an AR model is a factorization following the product rule, assuming each token is conditioned on all previous generated context.
Taking the L2R and R2L AR model as examples:
\begin{align}
    \mathcal{L}_{\text{L2R}} &= \sum_{i=1}^{N} \log P(y_i|y_{1:i-1}, X; \theta) \\
    \mathcal{L}_{\text{R2L}} &= \sum_{i=1}^{N} \log P(y_i|y_{i+1:N}, X; \theta)
\end{align}
where $X$ is the source text, $y_{1:i-1}$ and $y_{i+1:N}$ are previous outputs in opposite direction, $\theta$ is the learnable parameters, $N$ is the target length.

In the iterative-refinement based NAR model like CMLM \citep{DBLP:conf/emnlp/GhazvininejadLL19}, the conditional dependency is loosed, assuming the prediction of target token can be independent with each other, but conditioned on the output tokens (context) from last iteration:
\begin{equation}
  \mathcal{L}_{\text{CMLM}} = \sum_{y_i \in Y_{\text{mask}}^{(t)}} log P(y_i | X, Y_{\text{obs}}^{(t)}; \theta).  
\end{equation}
where $t$ is the iteration step $t=\{1,...,T\}$, $Y_{\text{obs}}$ are observable tokens (context), $Y_{\text{mask}} = Y \setminus Y_{\text{obs}}$ are masked tokens for predicting.
In each iteration, $N\frac{T-t}{T}$ of predicted tokens with low confidence will be re-masked and predicted again in the next iteration, conditioned on remaining high-confidence predictions as observable context until the last iteration.
At the initial iteration, the model determines the target length $N$ based on the source text $P(N|X)$ and makes the first step prediction with $N-2$ mask symbols as well as [BOS] and [EOS] input to the decoder, equivalent to merely conditioned on the source.

Instead of using the global context, in DisCo \citep{DBLP:conf/icml/KasaiCGG20}, the target token at each position is predicted with different context, namely, the disentangled context.
In such case, all tokens can be used for training and updated at each iteration during inference:
\begin{equation}
  \mathcal{L}_{\text{DisCo}} = \sum_{i=1}^{N} log P(y_i | X, Y_{\text{obs}}^{i,t}; \theta),  
\end{equation}
where $Y_{\text{obs}}^{i,t}$ is the context only for $y_i$.
The parallel easy-first decoding strategy is proposed (we call it easy first in following sections for simplicity) to improve the decoding efficiency, where the context of each token is composed by predictions at \textit{easier} positions determined in the first iteration:
\begin{equation}
  Y_{\text{obs}}^{i,t}=\{y_j^{t-1} | z(j)<z(i)\},  
\end{equation}
where $z(i)$ denotes the descending ordered rank of the probability $P_i$ computed in the first iteration.
During the training of CMLM and DisCo, a subset of tokens are selected as the context, CMLM updates parameters only with the loss on masked tokens while DisCO uses all tokens for updating.

In the Diformer, we aim to unify the two exclusive dependency assumptions \citep{DBLP:conf/nips/YangDYCSL19} of AR and NAR essentially by proposing a new training objective and model architecture that can make them trained jointly.

\subsection{Directional Language Model}
We aim to unify the AR and NAR decoding into three generation directions --- L2R, R2L and straight, i.e. making prediction on the target token at the rightward, leftward and the original position.
How to realize this goal is an open-question.
In this work, we achieve it by explicitly providing a direction instruction and corresponded contexts to the model.
Taking an example on the target sequence $Y=[A,B,C,D,E]$, the probability of $y_3 = C$ generated from three directions can be expressed as: $$P_3=
\begin{cases}
P(y_3=C | X, \{A,B\} )& \text{L2R}\\
P(y_3=C | X, \{D,E\} )& \text{R2L}\\
P(y_3=C | X, \{A,B,?,D,E\} )& \text{straight}
\end{cases}$$
where $?$ can be a mask symbol performing like a placeholder.

Formally, given the target sequence $Y=[y_1,...,y_N]$, token $y_i$ can be generated \textbf{from} direction $z_i \in \mathcal{Z}=\{R, S, L\}$ (i.e. L2R, straight and R2L) given the context $Y_{z_i}$ and X:
$$P(y_{z_i} | X, Y_{z_i}),$$
where $Y_{z_i}$ is determined by the direction $z_i$:
$$Y_{z_i}=
\begin{cases}
y_{1:i-1}& \ \ z_i=R\\
y_{i+1:N}& \ \ z_i=L\\
Y_{\text{obs}}^i& \ \ z_i=S
\end{cases}$$
When $z_i=R$ or $L$, the model works exactly same to the conventional AR model by conditioning on previously generated tokens at leftwards or rightwards.
When $z_i=S$, the model works in an iterative-refinement manner (e.g. mask-predict in CMLM or parallel easy-first in DisCO) by conditioning on a partially observed sequence $Y_{\text{obs}}^i$ with multiple tokens being masked including $y_i$, same as the disentangled context in DisCo.

We can thereby formulate the objective of directional language model as the expectation over all possible generation directions on each token:
\begin{equation}
    P(Y|X) = \mathbb{E}_{z_i \in \mathcal{Z}}\big{[}\prod_{i=1}^{N} P(y_{z_i} | X,Y_{z_i}) \big{]} \label{eq:DLM}
\end{equation}
The expectation can be approximated with sampling, similar to the permutation language model in \citep{DBLP:conf/nips/YangDYCSL19,DBLP:conf/coling/TianWCLZ20}, where a permutation of the sequence is sampled during training, we, instead, sample the direction \textbf{for each token}.
In this way, the factorization of DLM incorporates both conditional dependency assumption of AR, and conditional independence assumption of NAR, thereby makes the training objective closely related to the decoding methods.

\paragraph{Training}
The sampling of direction in DLM allows us to train the generation of all directions simultaneously, we introduce the detailed method in this section.

As we all know that the training of Transformer \citep{DBLP:conf/nips/VaswaniSPUJGKP17} can be paralleled with teacher forcing, achieved by feeding $y_{1:N-1}$ (context) to the model at once and computing the loss on $y_{2:N}$ (target).
The context and target sequence can be easily created by a shifting operation that aligns $y_{i-1}$ to $y_i$.

Diformer can also be trained in a similar way, but before that, we have to make a slight change when implementing the computation of the likelihood in Eq \ref{eq:DLM} due to the difficulty of creating the context sequence $Y_{z_i}$ with complicated dependencies.
The original equation aims to compute the likelihood on the ground-truth sequence $Y$ where each token is conditioned on a customized context determined by the sampled direction, meaning that the context sequence cannot be shared as Transformer does.
Creating specialized context for every token is non-trivial especially when encountered with position changing caused by the shifting when $z_i=R\ \text{or}\ L$.

For the convenience of the implementation, we fix the input sequence $y_{1:N}$ and create a new target sequence $Y^*$ where tokens are accordingly shifted with the sampled directions:
\begin{equation}
    P(Y^*|X) = \prod_{i=1}^N P(y_j | X, Y_{z_i}),
\end{equation}
where $j=i+1$ for $z_i=R$, $j=i-1$ for $z_i=L$ and $j=i$ for $z_i=S$.
When training on large corpus with random sampling on directions, we can say that $P(Y^*|X) \approx P(Y|X)$ theoretically.

Formally, let the source and target sequence as $X=[x_1,...,x_{|X|}]$ and $Y=[y_1,...,y_N]$ where $N$ is the target length.
Then, we uniformly sample a direction instruction sequence $Z=[z_1,...,z_N]$ with $N$ elements, where $z_1$ and $z_N$ are fixed to be $R$ and $L$ as they are [BOS] and [EOS], which can only be used to predict tokens inside the sequence for the AR setting, and can never be masked in the NAR setting.

The input sequence $Y_{\text{in}}$ is created by directly copying from ground-truth $Y$, which will be masked accordingly in the decoder to create the disentangled context.

According to the sampled direction sequence $Z$, we can now create the modified target sequence $Y^*$ by shifting tokens in $Y$ based on $z_i$, which is shown in Figure \ref{fig:diformer_model}.

To be compatible with the NAR decoding, we also predict the target length $P(N|X)$ with the same way as \citep{DBLP:conf/emnlp/GhazvininejadLL19}.
Note that the predicted length is only used for NAR decoding, the AR decoding still terminates when [EOS] or [BOS] is generated for L2R and R2L setting.

Finally, the cross-entropy loss is used for both generation ($\mathcal{L}_{\text{DLM}}$) and length prediction ($\mathcal{L}_{\text{LEN}}$) task, the overall loss can be obtained by adding them together:
\begin{equation}
    \mathcal{L}_{\text{Diformer}}=\mathcal{L}_{\text{DLM}} + \lambda\mathcal{L}_{\text{LEN}},
\end{equation}
where $\lambda$ is the factor on which the best performance can be obtained with the value of 0.1, after searched from 0.1 to 1.0 in the experiment.

\subsection{Directional Transformer}

\begin{figure}[t]
    \center
    \resizebox{0.8\columnwidth}{!}{
    \includegraphics{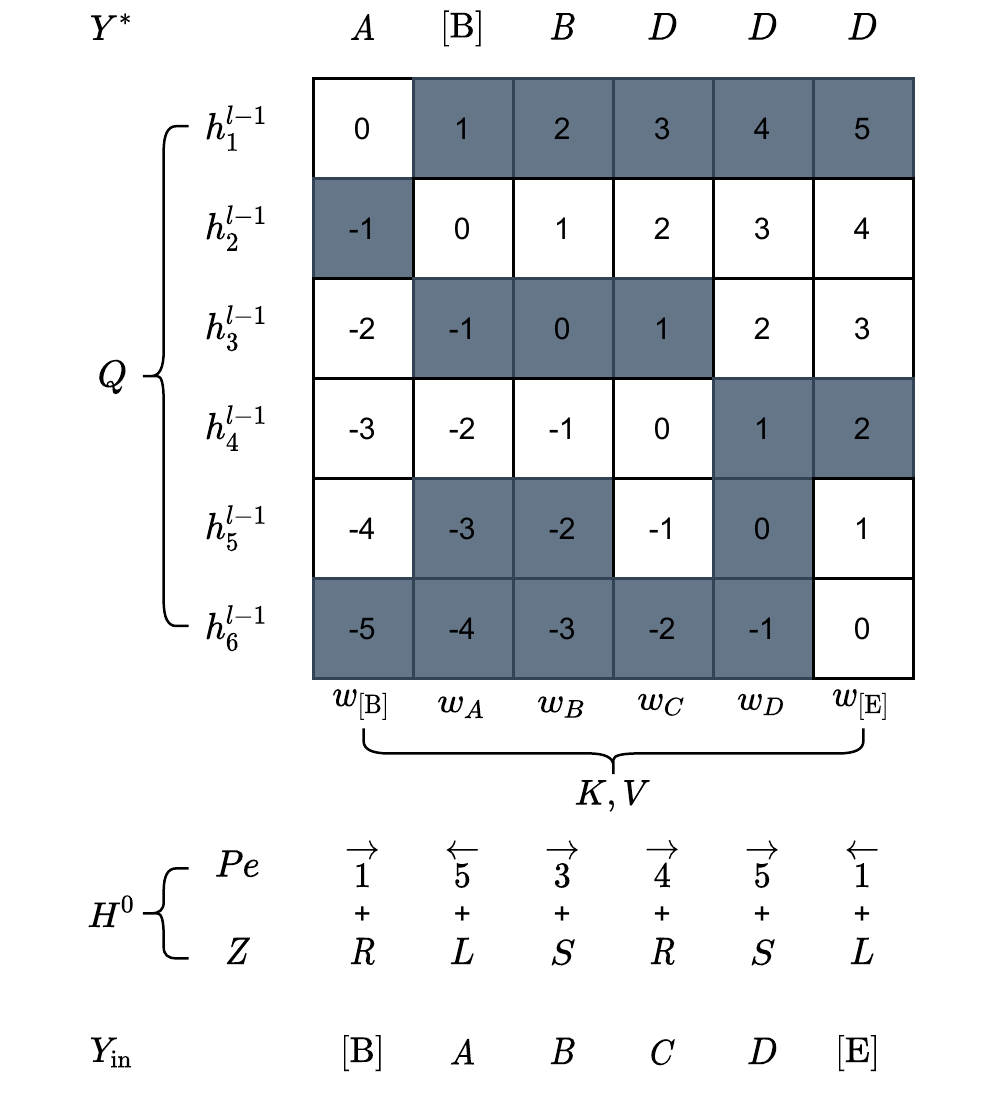}
    }
    \caption{An example of training Diformer with DLM, where values in grids are the relative distance of $K,V$ w.r.t $Q$, attention masks are indicated by dark grids.}
    \label{fig:diformer_model}
\end{figure}

\label{sec:model_arch}
Diformer is mainly built upon the Transformer \citep{DBLP:conf/nips/VaswaniSPUJGKP17} architecture but with several modifications for the compatible of the multi-directional generation, especially for avoiding the information leakage during training.

Specifically, we directly use the standard Transformer encoder in the Diformer, except that an additional MLP layer is added on top of it for length prediction.
For the decoder, several modifications are performed: 1) We introduce an additional embedding matrix to encode the direction instruction. 2) The original uni-directed positional embedding is expended to a bi-directed positional embedding. 3) We follow the work in DisCo to disentangle the context by de-contextualizing $K,V$ only with word embedding, and replacing the input of $Q$ in the first layer only with direction and position signal. 4) To compensate the removed positional information in $K,V$, we integrate the relative positional embedding in the self-attention, successfully resolved the problem on information leakage and the compatible of bi-directional generation. (see Appendix)

\paragraph{Directional Embeddings}
An embedding matrix is used to map the categorical variable $z_i$ into the hidden space, denoted as $\delta$, $\delta(z_i) \in \mathbb{R}^{d_\text{model}}$ where $d_\text{model}$ is hidden size of the model.
For simplicity, we directly use $z_i$ to represent the embedded direction at position $i$ in following sections.

The joint training of L2R and R2L can be problematic with the positional embedding of the original Transformer since the index is counted in a uni-directed order, which can be used for cheating under the bi-directional scenario since future positional index can leak information of the sentence length.

To solve this, we propose to make the positional embedding directed, achieved by encoding the position index counted oppositely based on the direction with separate parameters:
$$p_{z_i}=
\begin{cases}
\overrightarrow{Pe}(\overrightarrow{i})& \ \ z_i=R \ \ \text{or}\ \ z_i=S \\
\overleftarrow{Pe}(\overleftarrow{i})& \ \ z_i=L
\end{cases}$$
where $\overrightarrow{Pe}$ and $\overleftarrow{Pe}$ are different embedding matrics to encode position indices counting from L2R ($\overrightarrow{i}$) or R2L ($\overleftarrow{i}$) accordingly.
More detailed description can be found in Figure \ref{fig:diformer_model}.

Finally, we add encoded position and direction embeddings together as the initial hidden-state $h_i^{l=0}$ for the computation of $Q^{l=0}$ in the first self-attention layer $h_i^0 = p_{z_i} + z_i$:

\paragraph{Directional Self-Attention}
In DisCo, to prevent the information leakage from the disentangled context, the input representation for computing $K,V$ is de-contextualized by directly reusing the projection of input embeddings $k_i,v_i=\text{Proj}(w_i + p_i)$.
In Diformer, we have to further remove the positional information since the directed positional embedding can still be used for cheating in the computation of self-attention across layers.

Completely removing the the positional information on $K,V$ and only using the word-embedding $w_i$ can be harmful to the performance.
Therefore, we propose an alternative solution by replacing the removed absolute positional embedding with the relative positional embedding proposed in \citep{DBLP:conf/naacl/ShawUV18} for two reasons: 1) The relative position is computed in a 2 dimensional space, meaning that $p_{ij}$ and $p_{kj}$ for token $y_j$ is not shared between $y_i$ and $y_k$, which satisfies our requirements that each token in the context should have the position information only used for $y_i$ but not shared for $y_k$.
2) The position information is only injected during the computation of self-attention without affecting the original word embedding used in $K,V$.

Formally, we directly use the method in \citep{DBLP:conf/naacl/ShawUV18} but replace the hidden representation for computing $K,V$ with word embeddings:
\begin{align}
    h_i^{l'} &=\sum_{j=1}^{N} \alpha_{ij}(w_j W^V + p_{ij}^V) \\
    \alpha_{ij} &= \frac{\exp{e_{ij}}}{\sum_{k=1}^N \exp{e_{ik}}} \\
    e_{ij} &= \frac{h_i^{(l-1)} W^Q ( w_j W^K + p_{ij}^K)^{\top}}{\sqrt{d_{\text{head}}}}
\end{align}
where $h_i^{l'}$ is the output of the self-attention in current layer, $w_j$ is the word embedding, $p_{ij}^V, p_{ij}^K$ are embedded relative positions, $W^Q, W^K, W^V$ are parameters for $Q,K$ and $V$, $h_i^{l-1}$ is the last layer's hidden state, $d_\text{head}$ is the hidden size of a single head.
Two parameter matrics are used as embeddings --- $Re^K$ and $Re^V$, with the shape of $[2k+1, d_{\text{head}}]$, where $k$ is the max length.
$p_{ij}$ is obtained by embedding the distance between $i$ and $j$ clipped by the maximum length $k$.

Finally, a customized attention mask (see Figure \ref{fig:diformer_model}) is created during training to simulate specific dependencies based on the sampled direction sequence $Z$ with following rules:
\begin{itemize}
    \item If $z_i=R$, all tokens for $j > i$ will be masked.
    \item If $z_i=L$, all tokens for $j < i$ will be masked.
    \item If $z_i=S$, $y_i$ and a subset of randomly selected tokens will be masked following the method in \citep{DBLP:conf/icml/GhazvininejadKZ20}, excluding [BOS] and [EOS].
\end{itemize}

\paragraph{Inference}
Diformer can generate a sequence with 4 modes including L2R and R2L for AR decoding, mask-predict and parallel easy-first for NAR decoding.

For the AR decoding, the model works exactly same as the conventional Transformer, except that for each step, a fixed direction $z_i=R$ or $L$ should also be also be given, together with previously generated tokens, making it a pure-linear autoregressive generation.
Beam search can be directly used in both L2R and R2L decoding.
For the NAR decoding, the model uses mask-predict or easy-first by applying specific masking operation during each iteration, where all tokens are assigned with $z=S$.
Length beam can be used to further improve the performance.
Detailed examples are shown in the Appendix.

More importantly, we find that the multi-directional property of Diformer can be used for reranking, which is quite beneficial for the NAR decoding.
Specifically, compared to other NAR models that uses an external AR model for reranking, Diformer can do it all by its own without introducing additional computational costs.
For example, it first refines 5 candidates with 8 iterations and performs reranking with the rest of 2 iterations by re-using the encoder states and scoring candidates with L2R and R2L modes, which is equivalent to the computational cost of a 10-stepped refinement reported in CMLM.
The scores computed in two directions are averaged to obtain the final rank.
Experimental results show that 8 steps of refinement + 2 steps of reranking obtains significant performance improvements compared to 10 steps of refinement without re-ranking.
It can also be used for AR decoding, where all tokens are scored under the reversed direction, e.g. generating with L2R and scoring with R2L.
We name this method as \textbf{self-reranking}.

\section{Experiments}

\begin{table*}[t]
\centering
\resizebox{\textwidth}{!}{%
\begin{tabular}{@{}lcccccccc@{}}
\toprule
\multirow{2}{*}{} & \multicolumn{2}{c}{\textbf{En-De}} & \multicolumn{2}{c}{\textbf{De-En}} & \multicolumn{2}{c}{\textbf{En-Ro}} & \multicolumn{2}{c}{\textbf{Ro-En}} \\
 & \textbf{AR} & \textbf{NAR} & \textbf{AR} & \textbf{NAR} & \textbf{AR} & \textbf{NAR} & \textbf{AR} & \textbf{NAR} \\ \midrule
\multicolumn{9}{c}{\textbf{AR Models}} \\ \midrule
T-big \citep{DBLP:conf/nips/VaswaniSPUJGKP17} & 28.4 & - & - & - & - & - & - & - \\
T-base \citep{DBLP:conf/nips/VaswaniSPUJGKP17} & 27.3 & - & - & - & - & - & - & - \\
T-big (our impl, En$\leftrightarrow$De teacher) & 28.52 & - & 32.10 & - & - & - & - & - \\
T-base (our impl, En$\leftrightarrow$Ro teacher) & 27.67 & - & 31.12 & - & 35.29 & - & 34.02 & - \\
T-base + distill & 28.41 & - & 31.69 & - & 35.21 & - & 33.87 & - \\ \midrule
\multicolumn{9}{c}{\textbf{NAR models}} \\ \midrule
NAT \citep{DBLP:conf/iclr/Gu0XLS18} & - & 19.17 & - & 23.20 & - & 29.79 & - & 31.44 \\
iNAT \citep{DBLP:conf/emnlp/LeeMC18} & - & 21.61 & - & 25.48 & - & 29.32 & - & 30.19 \\
InsT \citep{DBLP:conf/icml/SternCKU19} & 27.29 & 27.41 & - & - & - & - & - & - \\
CMLM \citep{DBLP:conf/emnlp/GhazvininejadLL19} & - & 27.03 & - & 30.53 & - & 33.08 & - & 33.31 \\
LevT \citep{DBLP:conf/nips/GuWZ19} & - & 27.27 & - & - & - & - & - & 33.26 \\
DisCO \citep{DBLP:conf/icml/KasaiCGG20} & - & 27.34 & - & 31.31 & - & 33.22 & - & 33.25 \\ \midrule
\multicolumn{9}{c}{\textbf{Unified models}} \\ \midrule
\citet{mansimov2020generalized} & 25.66 & 24.53 & 30.58 & 28.63 & - & - & - & - \\
\citet{DBLP:conf/coling/TianWCLZ20} & 27.23 & 26.35 & - & - & - & - & - & - \\ \midrule
Diformer (ours) &  &  &  &  &  &  &  &  \\
- L2R & 28.35/\textbf{28.68} & - & 31.58/31.76 & - & 35.06/35.16 & - & 33.84/\textbf{33.92} & - \\
- R2L & 28.58/28.50 & - & \textbf{32.00}/31.78 & - & \textbf{35.17}/35.13 & - & 33.90/33.90 & - \\
- mask-predict & - & 27.51/\textbf{27.99} & - & 31.05/31.35 & - & 33.62/\textbf{34.37} & - & 32.68/33.11 \\
- easy-first & - & 27.35/27.84 & - & 31.21/\textbf{31.68} & - & 33.58/34.23 & - & 32.97/\textbf{33.34} \\ \bottomrule
\end{tabular}%
}
\caption{This table shows the overall performance of Diformer compared to the AR, NAR and unified models when decoding with AR or NAR strategies. T-big/-base is the abbreviation of Transformer-big/-base. The BLEU score using self-rerank (right) or not (left) is separated by /.}
\label{tab:perf}
\end{table*}

\subsection{Experimental Setup}

\paragraph{Data} We evaluate Diformer on 4 benchmarks including WMT14 En$\leftrightarrow$De (4.5M sentence pairs) and WMT16 En$\leftrightarrow$Ro (610k sentence pairs). 
The data is preprocessed in the same way with \citep{DBLP:conf/nips/VaswaniSPUJGKP17,DBLP:conf/emnlp/LeeMC18}, where each sentence is tokenized with Moses toolkit \citep{DBLP:conf/acl/KoehnHBCFBCSMZDBCH07} and encoded into subwords using BPE \citep{DBLP:conf/acl/SennrichHB16a}.
We follow \cite{DBLP:conf/iclr/Gu0XLS18,DBLP:conf/emnlp/GhazvininejadLL19,DBLP:conf/iclr/ZhouGN20} to create the knowledge distilled (KD) data with L2R Transformer-big and Transformer-base for En$\leftrightarrow$De and En$\leftrightarrow$Ro, the reported performance in the overall results are all obtained by training on the KD data.

\paragraph{Configuration} We follow the same configurations with previous works \citep{,DBLP:conf/nips/VaswaniSPUJGKP17,DBLP:conf/emnlp/GhazvininejadLL19,DBLP:conf/icml/GhazvininejadKZ20} on hyperparameters: $n_{(encoder + decoder)\_layers}$ = 6 + 6, $n_{heads}$ = 8, $d_{hidden}$ = 512, $d_{FFN}$ = 2048.
For customized components in Diformer, we tune the max relative distance $k$ in [1,8,16,256] and find that $k=256$ obtains best performance.
Adam \citep{DBLP:journals/corr/KingmaB14} is used for optimization with 128k tokens per batch on 8 V100 GPUs. 
The learning rate warms up for 10k steps to 5e-4 and decays with inversed-sqrt.
Models for En$\leftrightarrow$De and En$\leftrightarrow$Ro are trained for 300k and 100k steps, last 5 checkpoints are averaged for final evaluation.
We set beam size as 4 and 5 for AR and NAR decoding.
When decoding in NAR mode, we set the max iteration for mask-predict and easy-fist decoding as 10 without using any early-stopping strategy.
For fair comparison, we reduce the max iteration to 8 when decoding with self-reranking in NAR model.
Our model is implemented with pytorch \citep{torch} and fairseq \citep{DBLP:conf/naacl/OttEBFGNGA19}.
BLEU \citep{DBLP:conf/acl/PapineniRWZ02} is used for evaluation.

\subsection{Results \& Analysis}

We perform experiments on Diformer to evaluate its performance on three generation directions with four decoding strategies.
We mainly compare Diformer to three types of models: 1) the unified-models that is able to decode with multiple strategies, 2) pure AR model, i.e. standard Transformer, 3) pure NAR models. (see Table \ref{tab:perf})

\paragraph{Comparison with unified models} For the comparison to unified-models \citep{mansimov2020generalized,DBLP:conf/coling/TianWCLZ20}, Diformer outperforms others in all generation directions, by obtaining more than 1.5 BLEU.

As discussed in the section \ref{sec:intro}, their support on multiple generation strategies is achieved by applying certain position selection strategy on the masked language model or generating with certain permutation with the permutation language model.
This creates the gap between the training and inference since a specific decoding strategy might not be fully trained with the generalized objective as analyzed in \citep{mansimov2020generalized}.
So, compared to both, we use the task-specific modelling in exchange for better performance by abandon certain flexibility, thus makes the learned distribution to be same with the one used in decoding, which answers why Diformer performs better.

\paragraph{Comparison with AR models} For the En$\leftrightarrow$De dataset, since we use a larger teacher model (Transformer-big), therefore, we only compare Diformer with same sized Transformer-base trained on the raw and distilled data.
The Diformer outperforms Transformer trained on the raw data with a large margin and reaches the same level to the one trained on distilled data.
Interesting, the best performance of Diformer are usually obtained by the R2L decoding and the reranked results on L2R, the reason of it will be further discussed in ablation study sections.
For the En$\leftrightarrow$Ro dataset, Diformer can also obtain similar performance compared to the same sized Transformer trained on the distilled data produced by a same sized teacher.

\paragraph{Comparison with NAR models} Diformer is also competitive to a series of NAR models including iterative-refinement based and fully NAR models.
We speculate the strong performance of Diformer comes from the joint training of AR and NAR, since it is similar to the multi-task scenario, where tasks are closely correlated but not same.
This could be beneficial for the task that is more difficult i.e. NAR, because the learned common knowledge on AR tasks could be directly used in it.
By applying the self-reranking method, Diformer could obtain additional 0.5 BLEU over the strong baseline.

\subsection{Ablation Study}
In this section, we perform extra experiments to investigate factors that could influence the performance of Diformer and the mechanism behind it.
All experiments of ablation study are performed on the WMT14 En$\rightarrow$De dataset.

\paragraph{The influence of Knowledge Distillation}

\begin{table}[t]
\centering
\resizebox{\columnwidth}{!}{%
\begin{tabular}{@{}lcccc@{}}
\toprule
\textbf{Data Condition} & \textbf{R} & \textbf{L} & \textbf{mask-predict} & \textbf{easy-first} \\ \midrule
\textbf{T-base (our impl)} & \multicolumn{1}{l}{} & \multicolumn{1}{l}{} & \multicolumn{1}{l}{} & \multicolumn{1}{l}{} \\
Raw data & 27.67 & - & - & - \\
Distilled data & 28.41 & - & - & - \\ \midrule
\textbf{Diformer} &  &  &  &  \\
Raw data & 27.21 & 27.08 & 24.12 & 24.18 \\
Raw data (fixed right) & \multicolumn{1}{l}{27.63} & - & - & - \\
Distilled data & 28.35 & 28.55 & 27.51 & 27.35 \\ \bottomrule
\end{tabular}%
}
\caption{This table shows the performance of Transformer and Diformer trained on raw and distilled data where T-base represents for Transformer-base. An additional experiment with fixed $z_i=R$ for all tokens is also presented.}
\label{tab:kd_infl}
\end{table}

We train Diformer not only with distilled data but also with raw data as shown in table \ref{tab:kd_infl}.
The degradation of NAR decoding when training on raw data is not surprising which is a common problem faced by all NAR models.
However, the performance of AR decoding also degrades.
We speculate that on the raw data, the difficulty of learning to generate from straight and R2L increased significantly, making the model to allocate more capacity to fit them, resulting in the negative influence on the performance of L2R.
We verify this by fixing $z_i=R$ for all tokens and train the model on raw data.
The result confirms it because the performance recovers to its original level.
On the contrary, the knowledge distilled data is cleaner and more monotonous \citep{DBLP:conf/iclr/ZhouGN20}, making it easier to learn for all directions, and allows the model to allocate balanced capacity on each direction.
As for the better performance obtained by R2L decoding, we consider the reason is that, the R2L is able to learn the distilled data generated by the L2R teacher in a complementary manner, making it more efficient to learn the knowledge that cannot be learned by L2R due to the same modeling method.

\paragraph{The Importance of Relative Position}

\begin{table}[t]
\centering
\resizebox{0.8\columnwidth}{!}{%
\begin{tabular}{@{}lcccc@{}}
\toprule
\textbf{max $k$} & \textbf{R} & \textbf{L} & \textbf{mask-predict} & \textbf{easy-first} \\ \midrule
256 & 28.35 & 28.58 & 27.51 & 27.35 \\
16 & 28.51 & 28.48 & 27.25 & 27.32 \\
8 & 28.13 & 28.25 & 26.58 & 26.71 \\
1 & 26.81 & 26.85 & 18.78 & 19.53 \\ \bottomrule
\end{tabular}
}
\caption{This table shows the performance of Diformer with different max $k$.}
\end{table}

We also demonstrate the importance of the relative positional embedding by evaluating the model with different maximum relative distance $k$ and obtain the same conclusion \citep{DBLP:conf/naacl/ShawUV18} --- the distance should be at least 8.
Meanwhile, we observe that NAR is more sensitive to the positional information, which is reasonable, since the decoding of NAR is conditioned on the bi-directional context, where the positional information contains both distance and direction thereby is more important compared to that in AR.

\paragraph{Improvements of Self-Reranking}
\begin{figure}[t]
    \centering
    \resizebox{1\columnwidth}{!}{%
    \includegraphics{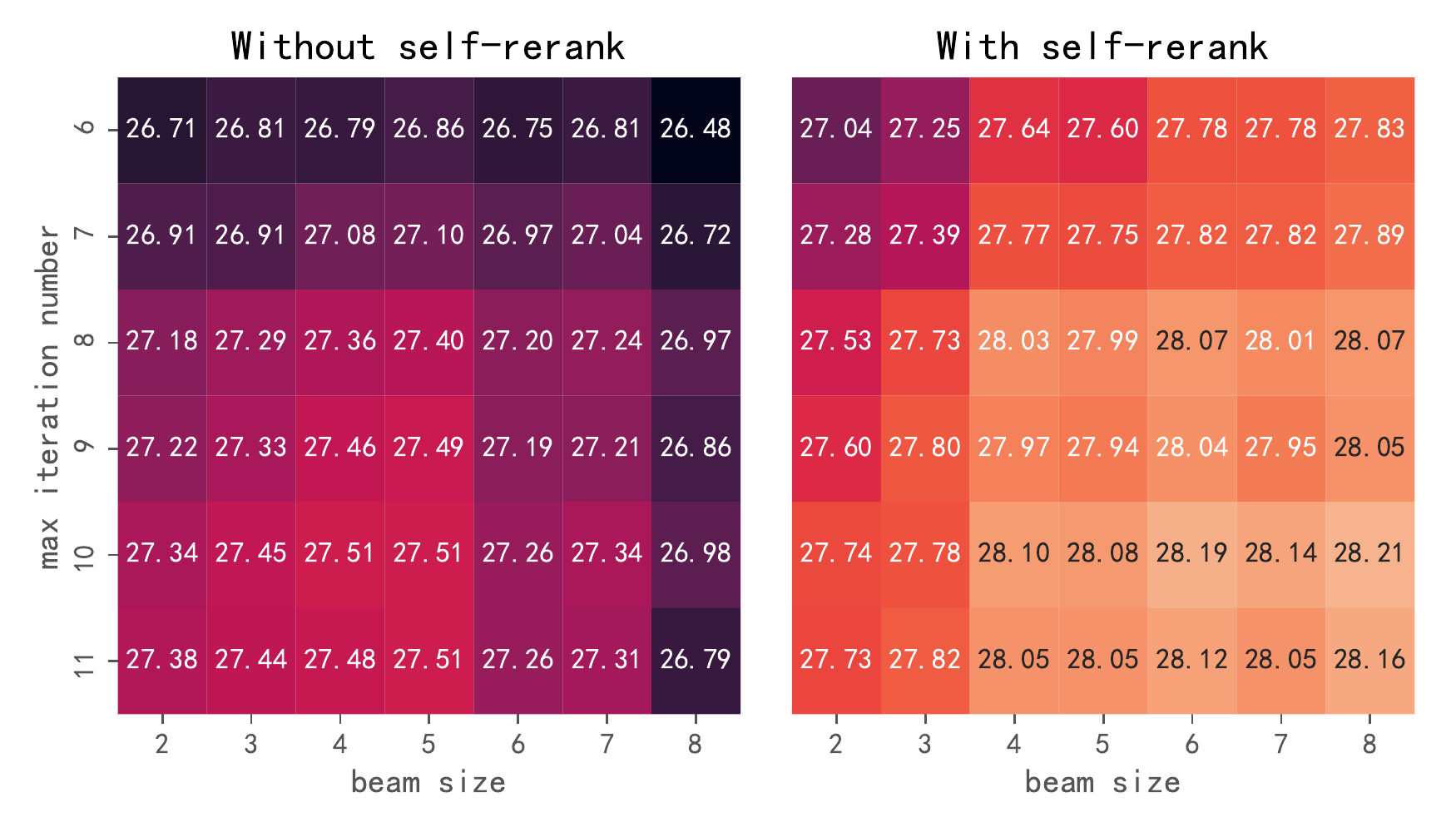}
    }
    \caption{The heatmap shows the BLEU score decoded with mask-predict when using self-reranking or not under different max iteration number and length beam size. }
    \label{fig:sr_perf}
\end{figure}
As shown in the overall results, self-reranking is a useful method to improve the performance especially for NAR decoding.
For the AR decoding, the improvements is not that significant since the outputs are already good enough for L2R or R2L, the tiny gap between reranking and generation direction cannot provide enough help, which indicates that using self-reranking for AR is not that profitable compared to NAR.

We further investigate its ability on NAR decoding (mask-predict) given different max iteration number and length beam size, as shown in Figure \ref{fig:sr_perf}.
It clearly shows that without reranking, the incorrect selection on beam candidates may even reduce the performance with larger beam size.
The use of self-reranking actually lets the performance and beam-size positively correlated, meaning that exchanging 2 steps of iteration with self-reranking can be profitable with larger beam size.
In practical usage of self-reranking, it is critical to find the optimal combination by balancing the beam size and max iteration number so that both high performance and low latency can be obtained.

\section{Conclusion}
In this paper, we present Directional Transformer which is able to model the autoregressive and non-autoregressive generation with a unified framework named Directional Language Model which essentially links two types of conditional language model with three generation directions.
Compared to previous works, Diformer exchanges the generalization on decoding strategies for better performance and thereby only support 4 decoding strategies.
Experimental results on WMT14 En$\leftrightarrow$De and WMT16 En$\leftrightarrow$Ro demonstrate that the unification of AR and NAR can be achieved by Diformer without losing any performance.
The bi-directional property of Diformer allows it to perform self-reranking which is especially useful for NAR decoding to improve performance with no additional computational cost.

Except from machine translation, Diformer can be easily extended to other tasks like language modeling by removing the dependency on $X$.
It has the potential to unify the representation learning and generation with a single model, which is actually our ongoing work.

\bibliography{acl2020}

\begin{thebibliography}{25}
\expandafter\ifx\csname natexlab\endcsname\relax\def\natexlab#1{#1}\fi

\bibitem[{Bahdanau et~al.(2015)Bahdanau, Cho, and
  Bengio}]{DBLP:journals/corr/BahdanauCB14}
Dzmitry Bahdanau, Kyunghyun Cho, and Yoshua Bengio. 2015.
\newblock \href {http://arxiv.org/abs/1409.0473} {Neural machine translation by
  jointly learning to align and translate}.
\newblock In \emph{3rd International Conference on Learning Representations,
  {ICLR} 2015, San Diego, CA, USA, May 7-9, 2015, Conference Track
  Proceedings}.

\bibitem[{Ghazvininejad et~al.(2020)Ghazvininejad, Karpukhin, Zettlemoyer, and
  Levy}]{DBLP:conf/icml/GhazvininejadKZ20}
Marjan Ghazvininejad, Vladimir Karpukhin, Luke Zettlemoyer, and Omer Levy.
  2020.
\newblock \href {http://proceedings.mlr.press/v119/ghazvininejad20a.html}
  {Aligned cross entropy for non-autoregressive machine translation}.
\newblock In \emph{Proceedings of the 37th International Conference on Machine
  Learning, {ICML} 2020, 13-18 July 2020, Virtual Event}, volume 119 of
  \emph{Proceedings of Machine Learning Research}, pages 3515--3523. {PMLR}.

\bibitem[{Ghazvininejad et~al.(2019)Ghazvininejad, Levy, Liu, and
  Zettlemoyer}]{DBLP:conf/emnlp/GhazvininejadLL19}
Marjan Ghazvininejad, Omer Levy, Yinhan Liu, and Luke Zettlemoyer. 2019.
\newblock \href {https://doi.org/10.18653/v1/D19-1633} {Mask-predict: Parallel
  decoding of conditional masked language models}.
\newblock In \emph{Proceedings of the 2019 Conference on Empirical Methods in
  Natural Language Processing and the 9th International Joint Conference on
  Natural Language Processing, {EMNLP-IJCNLP} 2019, Hong Kong, China, November
  3-7, 2019}, pages 6111--6120. Association for Computational Linguistics.

\bibitem[{Gu et~al.(2018)Gu, Bradbury, Xiong, Li, and
  Socher}]{DBLP:conf/iclr/Gu0XLS18}
Jiatao Gu, James Bradbury, Caiming Xiong, Victor O.~K. Li, and Richard Socher.
  2018.
\newblock \href {https://openreview.net/forum?id=B1l8BtlCb} {Non-autoregressive
  neural machine translation}.
\newblock In \emph{6th International Conference on Learning Representations,
  {ICLR} 2018, Vancouver, BC, Canada, April 30 - May 3, 2018, Conference Track
  Proceedings}. OpenReview.net.

\bibitem[{Gu et~al.(2019{\natexlab{a}})Gu, Liu, and
  Cho}]{DBLP:journals/tacl/GuLC19}
Jiatao Gu, Qi~Liu, and Kyunghyun Cho. 2019{\natexlab{a}}.
\newblock \href {https://transacl.org/ojs/index.php/tacl/article/view/1732}
  {Insertion-based decoding with automatically inferred generation order}.
\newblock \emph{Trans. Assoc. Comput. Linguistics}, 7:661--676.

\bibitem[{Gu et~al.(2019{\natexlab{b}})Gu, Wang, and
  Zhao}]{DBLP:conf/nips/GuWZ19}
Jiatao Gu, Changhan Wang, and Junbo Zhao. 2019{\natexlab{b}}.
\newblock \href {http://papers.nips.cc/paper/9297-levenshtein-transformer}
  {Levenshtein transformer}.
\newblock In \emph{Advances in Neural Information Processing Systems 32: Annual
  Conference on Neural Information Processing Systems 2019, NeurIPS 2019,
  December 8-14, 2019, Vancouver, BC, Canada}, pages 11179--11189.

\bibitem[{Kasai et~al.(2020)Kasai, Cross, Ghazvininejad, and
  Gu}]{DBLP:conf/icml/KasaiCGG20}
Jungo Kasai, James Cross, Marjan Ghazvininejad, and Jiatao Gu. 2020.
\newblock \href {http://proceedings.mlr.press/v119/kasai20a.html}
  {Non-autoregressive machine translation with disentangled context
  transformer}.
\newblock In \emph{Proceedings of the 37th International Conference on Machine
  Learning, {ICML} 2020, 13-18 July 2020, Virtual Event}, volume 119 of
  \emph{Proceedings of Machine Learning Research}, pages 5144--5155. {PMLR}.

\bibitem[{Kingma and Ba(2015)}]{DBLP:journals/corr/KingmaB14}
Diederik~P. Kingma and Jimmy Ba. 2015.
\newblock \href {http://arxiv.org/abs/1412.6980} {Adam: {A} method for
  stochastic optimization}.
\newblock In \emph{3rd International Conference on Learning Representations,
  {ICLR} 2015, San Diego, CA, USA, May 7-9, 2015, Conference Track
  Proceedings}.

\bibitem[{Koehn et~al.(2007)Koehn, Hoang, Birch, Callison{-}Burch, Federico,
  Bertoldi, Cowan, Shen, Moran, Zens, Dyer, Bojar, Constantin, and
  Herbst}]{DBLP:conf/acl/KoehnHBCFBCSMZDBCH07}
Philipp Koehn, Hieu Hoang, Alexandra Birch, Chris Callison{-}Burch, Marcello
  Federico, Nicola Bertoldi, Brooke Cowan, Wade Shen, Christine Moran, Richard
  Zens, Chris Dyer, Ondrej Bojar, Alexandra Constantin, and Evan Herbst. 2007.
\newblock \href {https://aclanthology.org/P07-2045/} {Moses: Open source
  toolkit for statistical machine translation}.
\newblock In \emph{{ACL} 2007, Proceedings of the 45th Annual Meeting of the
  Association for Computational Linguistics, June 23-30, 2007, Prague, Czech
  Republic}. The Association for Computational Linguistics.

\bibitem[{Lee et~al.(2018)Lee, Mansimov, and Cho}]{DBLP:conf/emnlp/LeeMC18}
Jason Lee, Elman Mansimov, and Kyunghyun Cho. 2018.
\newblock \href {https://doi.org/10.18653/v1/d18-1149} {Deterministic
  non-autoregressive neural sequence modeling by iterative refinement}.
\newblock In \emph{Proceedings of the 2018 Conference on Empirical Methods in
  Natural Language Processing, Brussels, Belgium, October 31 - November 4,
  2018}, pages 1173--1182. Association for Computational Linguistics.

\bibitem[{Mansimov et~al.(2020)Mansimov, Wang, Welleck, and
  Cho}]{mansimov2020generalized}
Elman Mansimov, Alex Wang, Sean Welleck, and Kyunghyun Cho. 2020.
\newblock \href {http://arxiv.org/abs/1905.12790} {A generalized framework of
  sequence generation with application to undirected sequence models}.

\bibitem[{Ott et~al.(2019)Ott, Edunov, Baevski, Fan, Gross, Ng, Grangier, and
  Auli}]{DBLP:conf/naacl/OttEBFGNGA19}
Myle Ott, Sergey Edunov, Alexei Baevski, Angela Fan, Sam Gross, Nathan Ng,
  David Grangier, and Michael Auli. 2019.
\newblock \href {https://doi.org/10.18653/v1/n19-4009} {fairseq: {A} fast,
  extensible toolkit for sequence modeling}.
\newblock In \emph{Proceedings of the 2019 Conference of the North American
  Chapter of the Association for Computational Linguistics: Human Language
  Technologies, {NAACL-HLT} 2019, Minneapolis, MN, USA, June 2-7, 2019,
  Demonstrations}, pages 48--53. Association for Computational Linguistics.

\bibitem[{Papineni et~al.(2002)Papineni, Roukos, Ward, and
  Zhu}]{DBLP:conf/acl/PapineniRWZ02}
Kishore Papineni, Salim Roukos, Todd Ward, and Wei{-}Jing Zhu. 2002.
\newblock \href {https://doi.org/10.3115/1073083.1073135} {Bleu: a method for
  automatic evaluation of machine translation}.
\newblock In \emph{Proceedings of the 40th Annual Meeting of the Association
  for Computational Linguistics, July 6-12, 2002, Philadelphia, PA, {USA}},
  pages 311--318. {ACL}.

\bibitem[{Paszke et~al.(2019)Paszke, Gross, Massa, Lerer, Bradbury, Chanan,
  Killeen, Lin, Gimelshein, Antiga, Desmaison, Kopf, Yang, DeVito, Raison,
  Tejani, Chilamkurthy, Steiner, Fang, Bai, and Chintala}]{torch}
Adam Paszke, Sam Gross, Francisco Massa, Adam Lerer, James Bradbury, Gregory
  Chanan, Trevor Killeen, Zeming Lin, Natalia Gimelshein, Luca Antiga, Alban
  Desmaison, Andreas Kopf, Edward Yang, Zachary DeVito, Martin Raison, Alykhan
  Tejani, Sasank Chilamkurthy, Benoit Steiner, Lu~Fang, Junjie Bai, and Soumith
  Chintala. 2019.
\newblock \href
  {http://papers.neurips.cc/paper/9015-pytorch-an-imperative-style-high-performance-deep-learning-library.pdf}
  {Pytorch: An imperative style, high-performance deep learning library}.
\newblock In \emph{Advances in Neural Information Processing Systems 32}, pages
  8024--8035. Curran Associates, Inc.

\bibitem[{Qi et~al.(2021)Qi, Gong, Jiao, Yan, Chen, Liu, Tang, Li, Chen, Zhang,
  Zhou, and Duan}]{DBLP:conf/icml/QiG0YCLTLCZ0D21}
Weizhen Qi, Yeyun Gong, Jian Jiao, Yu~Yan, Weizhu Chen, Dayiheng Liu, Kewen
  Tang, Houqiang Li, Jiusheng Chen, Ruofei Zhang, Ming Zhou, and Nan Duan.
  2021.
\newblock \href {http://proceedings.mlr.press/v139/qi21a.html} {{BANG:}
  bridging autoregressive and non-autoregressive generation with large scale
  pretraining}.
\newblock In \emph{Proceedings of the 38th International Conference on Machine
  Learning, {ICML} 2021, 18-24 July 2021, Virtual Event}, volume 139 of
  \emph{Proceedings of Machine Learning Research}, pages 8630--8639. {PMLR}.

\bibitem[{Sennrich et~al.(2016)Sennrich, Haddow, and
  Birch}]{DBLP:conf/acl/SennrichHB16a}
Rico Sennrich, Barry Haddow, and Alexandra Birch. 2016.
\newblock \href {https://doi.org/10.18653/v1/p16-1162} {Neural machine
  translation of rare words with subword units}.
\newblock In \emph{Proceedings of the 54th Annual Meeting of the Association
  for Computational Linguistics, {ACL} 2016, August 7-12, 2016, Berlin,
  Germany, Volume 1: Long Papers}. The Association for Computer Linguistics.

\bibitem[{Shaw et~al.(2018)Shaw, Uszkoreit, and
  Vaswani}]{DBLP:conf/naacl/ShawUV18}
Peter Shaw, Jakob Uszkoreit, and Ashish Vaswani. 2018.
\newblock \href {https://doi.org/10.18653/v1/n18-2074} {Self-attention with
  relative position representations}.
\newblock In \emph{Proceedings of the 2018 Conference of the North American
  Chapter of the Association for Computational Linguistics: Human Language
  Technologies, NAACL-HLT, New Orleans, Louisiana, USA, June 1-6, 2018, Volume
  2 (Short Papers)}, pages 464--468. Association for Computational Linguistics.

\bibitem[{Shu et~al.(2020)Shu, Lee, Nakayama, and
  Cho}]{DBLP:conf/aaai/ShuLNC20}
Raphael Shu, Jason Lee, Hideki Nakayama, and Kyunghyun Cho. 2020.
\newblock \href {https://aaai.org/ojs/index.php/AAAI/article/view/6413}
  {Latent-variable non-autoregressive neural machine translation with
  deterministic inference using a delta posterior}.
\newblock In \emph{The Thirty-Fourth {AAAI} Conference on Artificial
  Intelligence, {AAAI} 2020, The Thirty-Second Innovative Applications of
  Artificial Intelligence Conference, {IAAI} 2020, The Tenth {AAAI} Symposium
  on Educational Advances in Artificial Intelligence, {EAAI} 2020, New York,
  NY, USA, February 7-12, 2020}, pages 8846--8853. {AAAI} Press.

\bibitem[{Stern et~al.(2019)Stern, Chan, Kiros, and
  Uszkoreit}]{DBLP:conf/icml/SternCKU19}
Mitchell Stern, William Chan, Jamie Kiros, and Jakob Uszkoreit. 2019.
\newblock \href {http://proceedings.mlr.press/v97/stern19a.html} {Insertion
  transformer: Flexible sequence generation via insertion operations}.
\newblock In \emph{Proceedings of the 36th International Conference on Machine
  Learning, {ICML} 2019, 9-15 June 2019, Long Beach, California, {USA}},
  volume~97 of \emph{Proceedings of Machine Learning Research}, pages
  5976--5985. {PMLR}.

\bibitem[{Sutskever et~al.(2014)Sutskever, Vinyals, and
  Le}]{DBLP:conf/nips/SutskeverVL14}
Ilya Sutskever, Oriol Vinyals, and Quoc~V. Le. 2014.
\newblock \href
  {https://proceedings.neurips.cc/paper/2014/hash/a14ac55a4f27472c5d894ec1c3c743d2-Abstract.html}
  {Sequence to sequence learning with neural networks}.
\newblock In \emph{Advances in Neural Information Processing Systems 27: Annual
  Conference on Neural Information Processing Systems 2014, December 8-13 2014,
  Montreal, Quebec, Canada}, pages 3104--3112.

\bibitem[{Tian et~al.(2020)Tian, Wang, Cheng, Lian, and
  Zhang}]{DBLP:conf/coling/TianWCLZ20}
Chao Tian, Yifei Wang, Hao Cheng, Yijiang Lian, and Zhihua Zhang. 2020.
\newblock \href {https://doi.org/10.18653/v1/2020.coling-main.25} {Train once,
  and decode as you like}.
\newblock In \emph{Proceedings of the 28th International Conference on
  Computational Linguistics, {COLING} 2020, Barcelona, Spain (Online), December
  8-13, 2020}, pages 280--293. International Committee on Computational
  Linguistics.

\bibitem[{Vaswani et~al.(2017)Vaswani, Shazeer, Parmar, Uszkoreit, Jones,
  Gomez, Kaiser, and Polosukhin}]{DBLP:conf/nips/VaswaniSPUJGKP17}
Ashish Vaswani, Noam Shazeer, Niki Parmar, Jakob Uszkoreit, Llion Jones,
  Aidan~N. Gomez, Lukasz Kaiser, and Illia Polosukhin. 2017.
\newblock \href
  {https://proceedings.neurips.cc/paper/2017/hash/3f5ee243547dee91fbd053c1c4a845aa-Abstract.html}
  {Attention is all you need}.
\newblock In \emph{Advances in Neural Information Processing Systems 30: Annual
  Conference on Neural Information Processing Systems 2017, December 4-9, 2017,
  Long Beach, CA, {USA}}, pages 5998--6008.

\bibitem[{Welleck et~al.(2019)Welleck, Brantley, III, and
  Cho}]{DBLP:conf/icml/WelleckBDC19}
Sean Welleck, Kiant{\'{e}} Brantley, Hal~Daum{\'{e}} III, and Kyunghyun Cho.
  2019.
\newblock \href {http://proceedings.mlr.press/v97/welleck19a.html}
  {Non-monotonic sequential text generation}.
\newblock In \emph{Proceedings of the 36th International Conference on Machine
  Learning, {ICML} 2019, 9-15 June 2019, Long Beach, California, {USA}},
  volume~97 of \emph{Proceedings of Machine Learning Research}, pages
  6716--6726. {PMLR}.

\bibitem[{Yang et~al.(2019)Yang, Dai, Yang, Carbonell, Salakhutdinov, and
  Le}]{DBLP:conf/nips/YangDYCSL19}
Zhilin Yang, Zihang Dai, Yiming Yang, Jaime~G. Carbonell, Ruslan Salakhutdinov,
  and Quoc~V. Le. 2019.
\newblock \href
  {https://proceedings.neurips.cc/paper/2019/hash/dc6a7e655d7e5840e66733e9ee67cc69-Abstract.html}
  {Xlnet: Generalized autoregressive pretraining for language understanding}.
\newblock In \emph{Advances in Neural Information Processing Systems 32: Annual
  Conference on Neural Information Processing Systems 2019, NeurIPS 2019,
  December 8-14, 2019, Vancouver, BC, Canada}, pages 5754--5764.

\bibitem[{Zhou et~al.(2020)Zhou, Gu, and Neubig}]{DBLP:conf/iclr/ZhouGN20}
Chunting Zhou, Jiatao Gu, and Graham Neubig. 2020.
\newblock \href {https://openreview.net/forum?id=BygFVAEKDH} {Understanding
  knowledge distillation in non-autoregressive machine translation}.
\newblock In \emph{8th International Conference on Learning Representations,
  {ICLR} 2020, Addis Ababa, Ethiopia, April 26-30, 2020}. OpenReview.net.

\end{thebibliography}
\bibliographystyle{acl_natbib}

\appendix

\section{Appendices}
\label{sec:appendix}

\begin{figure*}[t]
    \centering
    \includegraphics[width=0.7\linewidth]{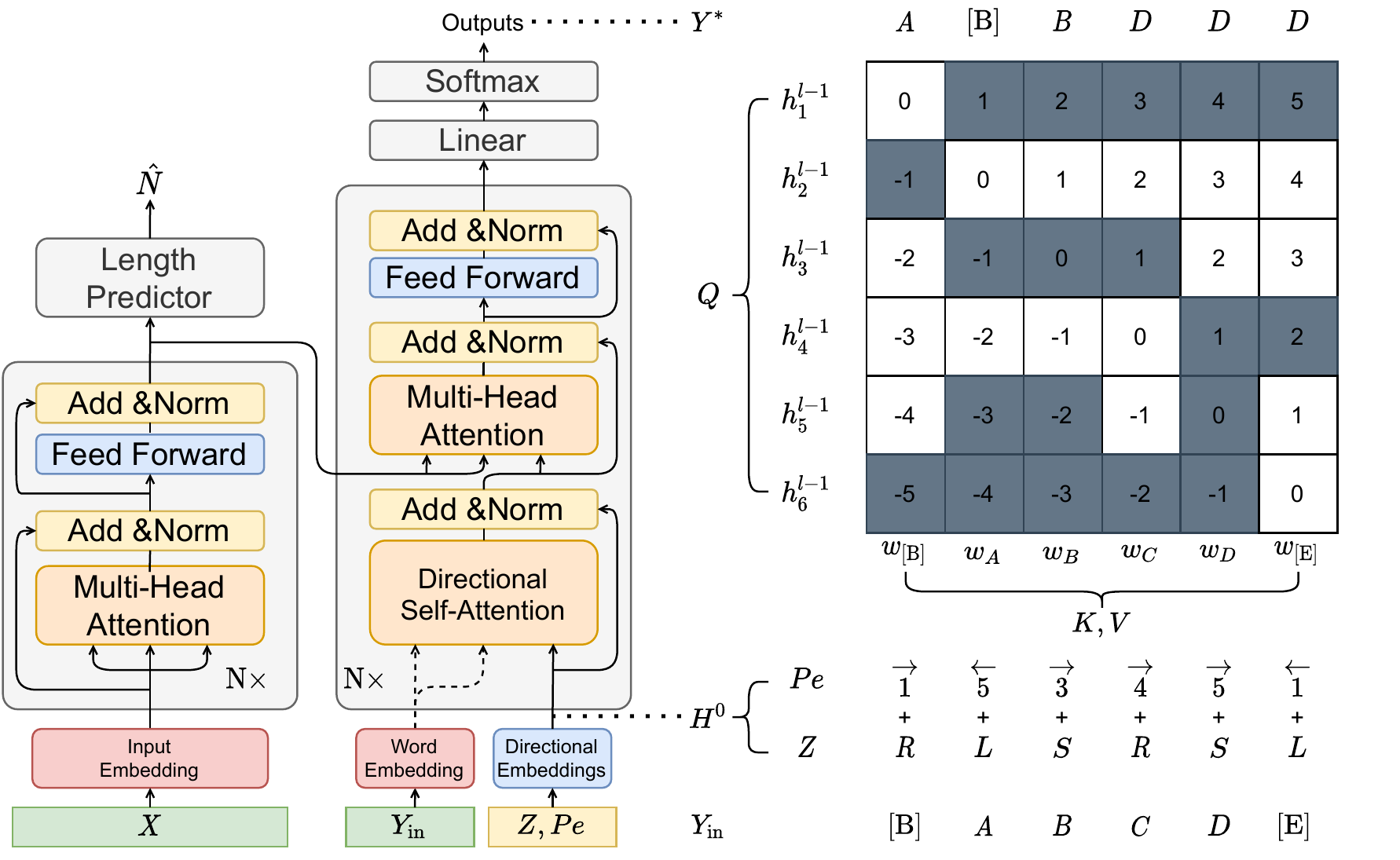}
    \caption{The architecture of the model (left) with the example of corresponded operations in key components during training (right).}
    \label{fig:my_label}
\end{figure*}

\begin{table*}[t]
\centering
\resizebox{\columnwidth}{!}{%
\begin{tabular}{@{}clrcc@{}}
\toprule
\textbf{Step} & \textbf{L2R} & \textbf{R2L} & \textbf{Mask-Predict} & \textbf{Easy-First} \\ \midrule
1 & \begin{tabular}[c]{@{}l@{}}A\\ {[}B{]}\\ R\end{tabular} & \begin{tabular}[c]{@{}r@{}}E\\ {[}E{]}\\ L\end{tabular} & \begin{tabular}[c]{@{}c@{}}{[}B{]} - - B - - {[}E{]}\\ {[}B{]} - - - - - {[}E{]}\\ S S S S S\end{tabular} & \begin{tabular}[c]{@{}c@{}}{[}B{]} A A C C D {[}E{]}\\ {[}B{]} - - - - - {[}E{]}\\ S S S S S\end{tabular} \\ \midrule
2 & \begin{tabular}[c]{@{}l@{}}A B\\ {[}B{]} A\\ R R\end{tabular} & \begin{tabular}[c]{@{}r@{}}D E\\ E {[}E{]}\\ L L\end{tabular} & \begin{tabular}[c]{@{}c@{}}{[}B{]} A B - - - {[}E{]}\\ {[}B{]} - - B - - {[}E{]}\\ S S S S S\end{tabular} & \begin{tabular}[c]{@{}c@{}}{[}B{]} A B C C E {[}E{]}\\ {[}B{]} A A C C D {[}E{]}\\ S S S S S\end{tabular} \\ \midrule
3 & \begin{tabular}[c]{@{}l@{}}A B C\\ {[}B{]} A B\\ R R R\end{tabular} & \begin{tabular}[c]{@{}r@{}}C D E\\ D E {[}E{]}\\ L L L\end{tabular} & \begin{tabular}[c]{@{}c@{}}{[}B{]} A B - - E {[}E{]}\\ {[}B{]} A B - - - {[}E{]}\\ S S S S S\end{tabular} & \begin{tabular}[c]{@{}c@{}}{[}B{]} A B C D E {[}E{]}\\ {[}B{]} A B C C E {[}E{]}\\ S S S S S\end{tabular} \\ \midrule
4 & \begin{tabular}[c]{@{}l@{}}A B C D\\ {[}B{]} A B C\\ R R R R\end{tabular} & \begin{tabular}[c]{@{}r@{}}B C D E\\ C D E {[}E{]}\\ L L L L\end{tabular} & \begin{tabular}[c]{@{}c@{}}{[}B{]} A B C - E {[}E{]}\\ {[}B{]} A B - - E {[}E{]}\\ S S S S S\end{tabular} & - \\ \midrule
5 & \begin{tabular}[c]{@{}l@{}}A B C D E\\ {[}B{]} A B C D\\ R R R R R\end{tabular} & \begin{tabular}[c]{@{}r@{}}A B C D E\\ B C D E {[}E{]}\\ L L L L L\end{tabular} & \begin{tabular}[c]{@{}c@{}}{[}B{]} A B C D E {[}E{]}\\ {[}B{]} A B C - E {[}E{]}\\ S S S S S\end{tabular} & - \\ \midrule
6 & \begin{tabular}[c]{@{}l@{}}A B C D E {[}E{]}\\ {[}B{]} A B C D E\\ R R R R R R\end{tabular} & \begin{tabular}[c]{@{}r@{}}{[}B{]} A B C D E\\ A B C D E {[}E{]}\\ L L L L L L\end{tabular} & - & - \\ \bottomrule
\end{tabular}%
}
\caption{An example of 4 decoding strategies with the output, context input and direction input placed in three sub-rows, where [B] and [E] represent for [BOS] and [EOS] token, - represents for mask.}
\label{tab:inference}
\end{table*}

\end{document}